\documentclass{article}
\usepackage{graphicx}
 \usepackage[preprint]{neurips_2026}

\usepackage[utf8]{inputenc} 
\usepackage[T1]{fontenc}    
\usepackage{hyperref}       
\usepackage{url}            
\usepackage{booktabs}       
\usepackage{amsfonts}       
\usepackage{nicefrac}       
\usepackage{microtype}      
\usepackage{xcolor}         

\title{ILM: An AI-Powered Storytelling Educational Tool}

\author{%
  Suhaila Mohammed \qquad Abdelaziz Serour \qquad Allison Lahnala \\
  Department of Computing and Software, Faculty of Engineering \\
  McMaster University, Hamilton, ON, Canada \\
  \texttt{\{suhai1, seroura, lahnalaa\}@mcmaster.ca}
}

\begin{document}

\maketitle

\begin{abstract}
Digital technologies have made Islamic narratives more accessible, but existing platforms provide limited support for structured learning and comprehension of these stories, particularly in Arabic and multilingual settings. We present ILM, an interactive educational platform for Stories of the Prophets that combines Arabic natural language processing, structured knowledge representation, and retrieval-based question generation. Admin-approved Arabic narratives are processed by a Knowledge Graph (KG) Constructor Engine that identifies entities and narrative relationships and stores them as structured knowledge, enabling learners to explore stories through a visual story map and answer entity- and relation-based questions generated from the KG. Separately, a multilingual retrieval pipeline retrieves relevant passages from the original narratives to generate multiple-choice and open-ended comprehension questions. For open-ended questions, an LLM-as-a-Judge evaluates learners' answers against the retrieved passages and reference answers to determine correctness. The platform also incorporates Quranic content as a separate enrichment layer, allowing selected narratives to be supplemented with source-supported information. By combining structured knowledge with passage-based retrieval, ILM supports narrative exploration, comprehension, and assessment across Arabic and multilingual content. The system demonstrates the feasibility of combining structured knowledge representation and retrieval-based generation to support interactive learning of Islamic narratives. A demo is available at \href{https://anonymous.4open.science/r/mml-5FCF/}{anonymous.4open.science/r/mml-5FCF}.
\end{abstract}

\section{Introduction}
Knowledge, or ilm, holds an important position in Islam and is highly valued by Muslims. Islamic knowledge has historically been passed from one generation to the next through teaching, study, narration, and direct engagement between teachers and learners. Narration has an important role in Islamic knowledge transmission, particularly through a framework and method called \textit{qaṣaṣ} (Qur'anic narratives). The Qur'an uses narrative to convey guidance and lessons, highlighting its pedagogical significance within Islamic education \cite{yusoff2023narrative}.

In today's world, the way knowledge is transmitted has changed significantly with the rise of digital technologies where information is now more widely available and accessible than before. On the other hand, digital platforms have become an increasingly common medium through which people learn. However, the availability of information through digital platforms does not necessarily ensure meaningful engagement or learning. Recent studies have shown that digital storytelling of prophetic narratives can support moral and religious character formation, as well as learner motivation and engagement, particularly when stories are well designed and appropriately mediated by adults \cite{aziz2026digital,prastiwi2024digitalization,rohman2025learning}.

Therefore, we developed a storytelling platform for children and young adults that combines reading, visualization, and interactive learning. We begin by focusing on prophetic stories. Through an interactive story map, users can visualize key entities in each story, such as the prophets and other people, locations, and important objects, and explore the relationships between them. They can also evaluate their understanding of the stories through quizzes that assess both factual knowledge and story comprehension.

\section{Related Work}
\paragraph{Existing apps and AI systems:} A range of apps and AI systems have risen to support the learning and exploration of prophetic and Islamic stories. Apps such as Qissah\cite{qissah2025app} and Qalam\cite{qalam2025app} provide 40+ narrated Qur’an and prophet stories with professional voiceover and multilingual support, while platforms such as Thurayya\cite{thurayya2025app}, Miraj Stories\cite{miraj2025app}, Quran Stories 4 Kids \cite{quranstories4kids2026}, and Kisah Nabi \cite{kisahnabi2025app} combine animated or interactive stories, audio narration, and games to present sīrah and qiṣaṣ al-anbiyā' in engaging formats. Conversational AI has also begun to be used in Islamic education, with general-purpose systems such as ChatGPT \cite{openai2026chatgpt} and specialized systems such as MuslimGPT \cite{muslimgpt2026app}, TheoAI \cite{theoai2026web}, providing sourced answers grounded in the Qur’an and authenticated hadith.

\paragraph{Knowledge graph and RAG systems:}
Knowledge graphs (KGs) have been used in Islamic texts to structure entities and relations in Qur’anic and hadith corpora for semantic search and question answering \cite{semanticHadithKG2026, hadithKGQA2022, ijasce2025ner}. More generally, narrative-centric KGs represent events, entities, and temporal or causal links to support story understanding and interactive exploration \cite{narrativeKG2026, inaSurvey2026, hierarchicalKGComics2026}. Retrieval-augmented generation (RAG) frameworks further combine LLMs with retrieved context, with recent work exploring temporally coherent passage assembly for narrative QA \cite{chronorag2026}.

Despite these developments, most work remains English-focused, with limited support for Arabic and educator control. We address these gaps through Arabic- and English-based learning that enables learners to explore narrative structure and relationships, assess factual and comprehension-based knowledge, receive story-grounded feedback, and learn from educator-controlled content.

\section{System Architecture and Design}
\begin{figure}[t]
    \centering
    \includegraphics[width=\linewidth]{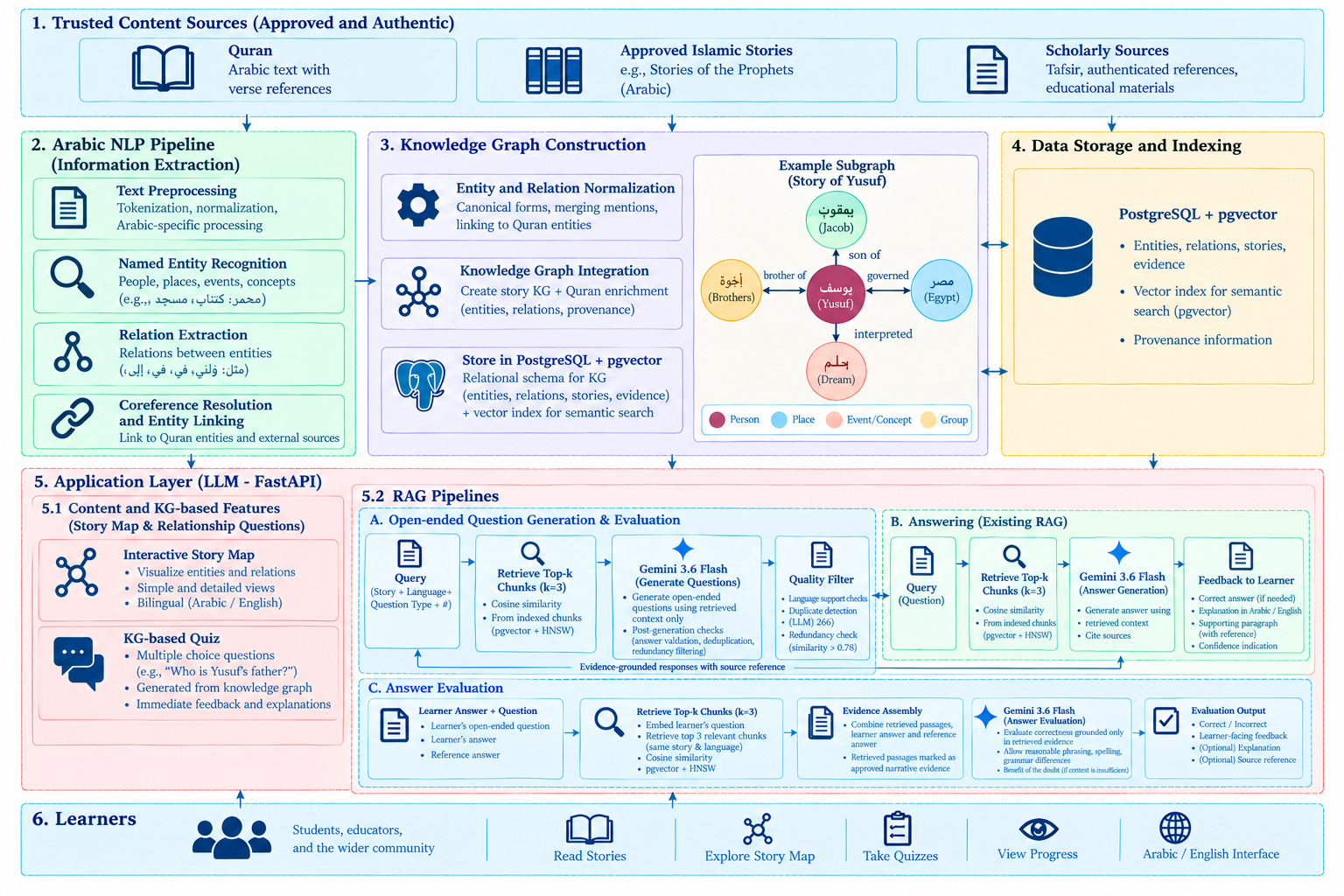}
    \caption{Overall architecture of the proposed ILM platform.}
    \label{fig:system_architecture}
\end{figure}

The platform comprises end-user and administrative web portals built with React.js and connected to a FastAPI backend. PostgreSQL serves as the primary data store, with pgvector supporting vector storage and similarity search. The backend integrates a KG Constructor Engine (KG Engine) for structured knowledge extraction from Arabic narratives approved by educators with admin privileges, and a multilingual retrieval pipeline for question generation and answer evaluation. Figure~\ref{fig:system_architecture} presents the overall architecture.

\subsection{Knowledge Graph Constructor Engine}
The KG Engine constructs a story-specific KG from Arabic paragraphs through named entity recognition (NER), story-scoped entity linking, candidate pair generation, relation extraction, and semantic resolution. Extracted entities and relations are converted into structured KG records.

Each paragraph is processed using a four-expert nested NER ensemble comprising AraBERTV2 \cite{antoun-etal-2020-arabert}, CAMELBERT-Mix\cite{inoue-etal-2021-interplay}, MARBERTV2\cite{abdul-mageed-etal-2021-arbert}, and an AraBERTv2 span-based expert. An entity span is retained when at least two experts produce an exact-span match. XLM-R\cite{conneau-etal-2020-unsupervised} is additionally used as a complementary safe-addition component for mentions not sufficiently supported by the primary ensemble. The resulting mentions are passed to a story-scoped Islamic entity linker, which maps surface forms, aliases, titles, and pronouns to canonical entities within the story.

After entity linking, candidate entity pairs are passed to a two-stage relation extraction pipeline. An AraBERT-based binary classifier first identifies related pairs, followed by a relation classifier using the WOJOOD relation scheme\cite{aljabari-etal-2025-wojoodrelations}. The top-5 relations are then evaluated by an NLI model using the original text as the premise to resolve semantic ambiguities, such as \textit{birth place} versus \textit{lives in}. Selected relation types are further refined by relation-specific resolvers for family, prophetic mission, and major events before being added to the knowledge graph.

The extraction pipeline operates over the broader WOJOOD ontology\cite{aljabari-etal-2025-wojoodontology}, while the educational application exposes nine controlled predicates: \textit{child\_of}, \textit{father\_of}, \textit{sibling\_of}, \textit{sent\_to}, \textit{thrown\_into}, \textit{taken\_to}, \textit{imprisoned\_in}, \textit{appointed\_over}, and \textit{reunited\_with}. Although the KG Engine extracts relations from the full WOJOOD ontology\cite{aljabari-etal-2025-wojoodontology}, only nine selected relations are used in the educational application for learner activities and deterministic assessment.

The extracted entities, mentions, and relations are converted into structured knowledge graph records before being stored in database. Each relation is validated against its entity endpoints and stored together with information about its source and extraction process. This includes the supporting paragraph, source information, original relation prediction, and, where applicable, the semantic resolution rule used to obtain the final relation. The graph is maintained in synchronization with the approved narrative content. When a paragraph is modified, the KG records previously associated with that paragraph are retracted and the updated paragraph is processed again.

A separate Quran enrichment layer is derived from the \textit{Tanzil Quran Text--Simple Clean v1.1} (CC BY 3.0) \cite{tanzil2021quran}. Relations are manually verified against the Quranic source and stored with their corresponding verses. This layer adds relations such as \texttt{interpreted\_for}, \texttt{tempted}, and \texttt{summoned}, but remains separate from the nine-predicate educational ontology and is excluded from deterministic quiz generation. Entity representation is conservative and limited to information explicitly supported by the source, distinguishing verified Quranic knowledge from automatically extracted narrative knowledge.

\subsection{Multilingual Retrieval and RAG Pipeline}
The retrieval pipeline operates independently of the KG Engine and supports passage-based open-ended question generation and evaluation. Each approved narration is divided into paragraphs, with approved translations maintained across languages. Following human review, each paragraph translation is indexed as a retrieval chunk, preserving a direct mapping between retrieved content and its source paragraph. Chunks are embedded using Cohere's \texttt{embed-multilingual-v3.0} model\cite{cohere2024embedmultilingual}, producing 1024-dimensional representations stored in PostgreSQL with pgvector and an HNSW index for efficient cosine-similarity search. Arabic text is normalized before embedding and retrieval to reduce surface-form variation.

For question generation, the administrator specifies the story, language, question type, and number of questions. The query is embedded and matched against the chunk embeddings, with the top $k=3$ chunks provided to Gemini 3.6 Flash\cite{google2026gemini} as source context. The model is instructed\footnote{See Appendix~\ref{app:open-ended-prompt}.} to generate questions solely from the retrieved passages. Post-generation checks verify answer support against the source passages, detect within-chunk duplicates using Jaccard word overlap ($\geq 0.6$), and flag potentially redundant questions across the story using embedding similarity ($>0.78$) for manual review. Duplicate generation is retried once before the question is discarded.

\subsection{Question Evaluation}

The end-user portal combines two assessment mechanisms: knowledge-graph-based questions and retrieval-based open-ended questions. After completing all paragraphs of a story, the learner can attempt the corresponding quiz. The two question types use different evaluation strategies because they rely on different representations of the underlying knowledge.

\subsubsection{Knowledge-Graph-Based Evaluation}
Knowledge-graph-based questions use the story knowledge graph as the source of truth. Questions are generated from graph triples representing entities and relations, with each question linked to its source triple and expected answer. This enables deterministic evaluation without a language model: answers are normalized and matched against the canonical graph value and registered aliases, while multiple-choice answers are determined directly from the corresponding graph entity or relation.

\subsubsection{Retrieval-Based Answer Evaluation}

For retrieval-based open-ended questions, the learner's question is embedded and used to retrieve the top $k=3$ relevant chunks from the same story and language. The retrieved passages, together with the learner's answer and the reference answer, are provided to Gemini\cite{google2026gemini} for evaluation. The retrieved passages serve as the evidence available to the evaluator, ensuring that the assessment remains grounded in the approved narrative content. Gemini\cite{google2026gemini} determines whether the learner's answer is correct based on the retrieved evidence while allowing for reasonable differences in phrasing, spelling, and grammar. When the retrieved passages do not contain sufficient information to establish that an answer is incorrect, the system follows a benefit-of-the-doubt policy rather than penalizing the learner based on information unavailable in the retrieved context. The model returns a correct or incorrect and learner-facing feedback.

\section{Conclusion}
We presented ILM, an interactive educational platform that combines structured knowledge representation and multilingual retrieval to support learning from Islamic narratives. ILM establishes a framework for connecting narrative content with interactive exploration and comprehension assessment while maintaining links to the underlying source material. This work provides a foundation for extending Arabic and multilingual educational applications and for future empirical evaluation with learners and educators. The platform can support accessible and interactive learning of Islamic narratives. However, automatically extracted relations and generated questions may contain errors, which we mitigate through approved narrative content, human review, and source-verified Quranic enrichment. A demonstration video showcasing the ILM platform and its interactive learning capabilities is available here:
\textbf{\href{https://anonymous.4open.science/r/mml-5FCF/}{demo video}}.

\section{Future Work}
Several directions can further extend the platform. First, we plan to expand the story corpus to include a wider range of prophetic narratives and evaluate the system across different texts, as the current study is limited by its evaluation on a single story. Second, we aim to conduct formal empirical studies with children and educators to measure the efficacy of the platform in improving narrative comprehension, engagement, and knowledge retention. Third, we could also further improve engagement by incorporating multimedia such as GIFs, animations, and short videos, along with more interactive assessment activities such as matching characters and relationships, ordering events, and completing story maps. 

\begin{ack}
Computing and API costs were supported by a start-up fund awarded to Allison Lahnala by McMaster University. The authors declare no competing interests. The views expressed are those of the authors and do not necessarily reflect the views of McMaster University.\end{ack}

\bibliographystyle{plain}
\bibliography{reference}







\appendix

\section{Gemini 3.6 Flash Prompts}

\subsection{Open-Ended Question Generation {Prompt}}
\label{app:open-ended-prompt}
You write reading-comprehension questions for children aged \texttt{{age}}.

Use ONLY the SOURCE provided below. Do not add names, events, places, or details that are not present in the SOURCE. If the SOURCE does not provide enough information to answer a question, do not ask about it.

Write ONE open-ended question that requires the child to answer in their own words, using one or two sentences. The question must be answerable from the SOURCE alone.

Ask about the story itself, not about the writing or the source. Do not use phrases such as according to the text'' or the story states.''

Also provide the expected answer as one short sentence that a child might reasonably write.

Write everything in \texttt{{language}}.

\texttt{{avoid}}

Return only valid JSON in the following format:

\begin{verbatim}
{"prompt": "...", "reference_answer": "...", "explanation": "..."}
\end{verbatim}

\subsection{Open-Answer Judge Prompt}
A child aged \texttt{{age}} was asked a question about a story and wrote an answer.

Decide whether the child's answer correctly answers the QUESTION using ONLY the PASSAGES provided below. Do not use any outside knowledge or information about the story.

If the PASSAGES do not provide enough information to determine whether the answer is correct, give the child the benefit of the doubt and mark the answer as correct.

Be generous about wording, spelling, grammar, and sentence structure. The child does not need to use the same words as the expected answer. Mark an answer as correct when its meaning is consistent with the PASSAGES and it demonstrates understanding of what the question asks.

Mark an answer as incorrect only when:
\begin{itemize}
\item it contradicts information in the PASSAGES;
\item it gives information that is incompatible with the PASSAGES; or
\item it fails to answer the main point of the QUESTION.
\end{itemize}

Do not require every detail from the reference answer if the child's answer is otherwise correct. Do not mark an answer wrong merely because it is incomplete in a minor way, uses different wording, or contains harmless spelling or grammar mistakes.

Write feedback for the child in \texttt{{language}}, using one short, encouraging sentence. If the answer is correct, briefly say what they got right. If the answer is incorrect, kindly provide the correct answer based only on the PASSAGES. Never mention passages, sources, texts, evaluation, or judging.

\begin{verbatim}
QUESTION: {question}
THEIR ANSWER: {answer}
{reference}
PASSAGES:
{passages}
{avoid}
Return only valid JSON:
{"is_correct": true, "feedback": "..."}
\end{verbatim}

\subsection{Shared Block for Question Generation Prompt}

\textbf{Shared injected block (\texttt{{avoid}})}

These questions have already been asked about this same passage. Ask about something else in it, a different detail, person, action, or consequence. Do not rephrase or repeat any of them:
\begin{verbatim}
{existing}
\end{verbatim}

\end{document}